\documentclass[letterpaper, 10 pt, conference]{ieeeconf}  % Comment this line out
\usepackage{graphicx}
\IEEEoverridecommandlockouts                              % This command is only
\usepackage[utf8]{inputenc}
\usepackage[T1]{fontenc}

\title{\LARGE \bf
Residual Quantity in Percentage of Factory Machines Using Computer Vision and Mathematical Methods
}

\author{Seunghyeon Kim$^{1}$, Jihoon Ryoo$^{2}$, Dongyeob Lee$^{3}$, and Youngho Kim$^{4}$% <-this % stops a space
}

\begin{document}

\maketitle
\thispagestyle{empty}
\pagestyle{empty}
%%%%%%%%%%%%%%%%%%%%%%%%%%%%%%%%%%%%%%%%%%%%%%%%%%%%%%%%%%%%%%%%%%%%%%%%%%%%%%%%
\section{INTRODUCTION}

\begin{figure}[htp]
    \centering
    \includegraphics[scale = 0.1]{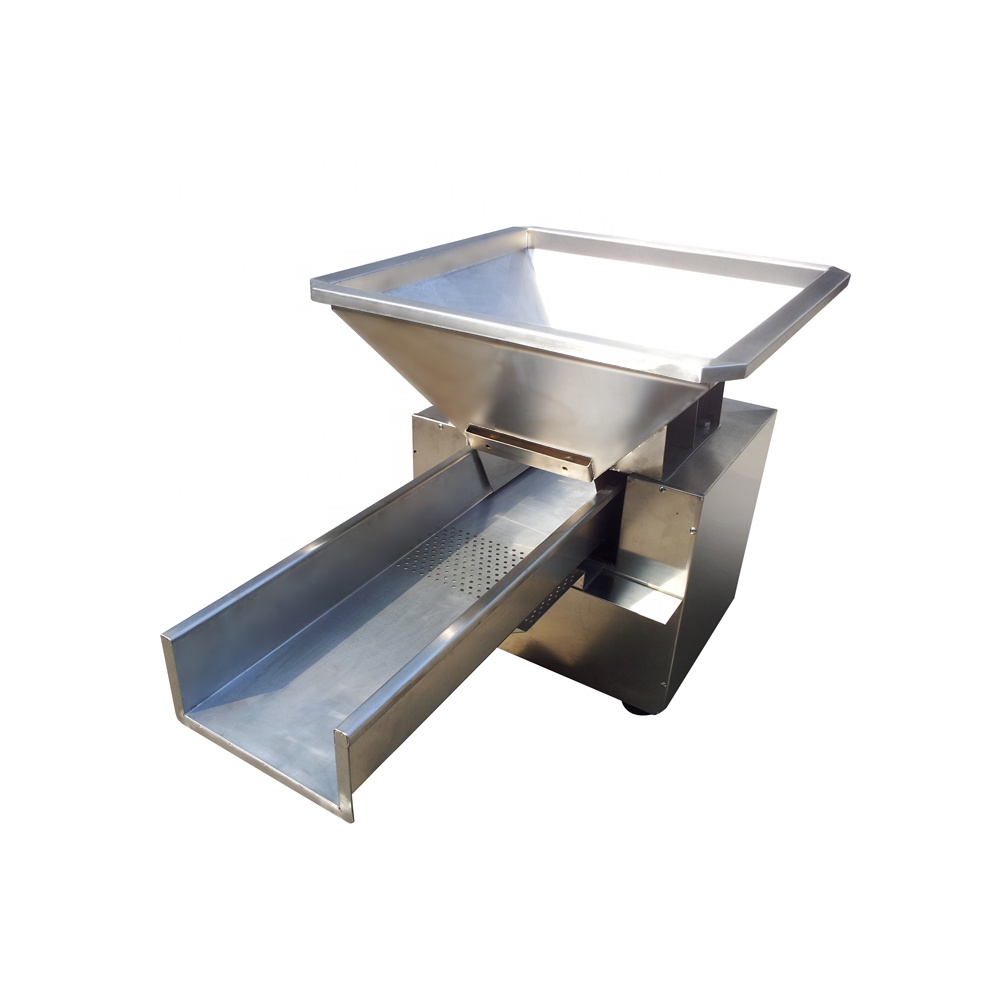}
    \caption{Image of a vibrating hopper [3]}
    \label{fig:hopper}
\end{figure}

Computer vision is an advanced computer science technique that has provided the basis for many new patents and APIs that have informed a multitude of applications. Computer vision involves all analyses, including machine learning, performed using pixels of images. Some manufacturing industries have recently discovered that they need human resources to monitor whether the machine is working properly and if adequate materials are entering the hopper. From an economic perspective, using humans to monitor resources manually can be an inefficient allocation of resources associated with welfare decline. Furthermore, the hopper is 2m tall, introducing the possibility of falls as personnel ascends a ladder for monitoring. A fall from the ladder could lead to an industrial accident and a significant loss of profit for the company. 

Using computer vision could eliminate these risks. While it may be possible to use the loadcell specific to the hopper to monitor the change in mass, such a solution could be very useful when the hopper does not vibrate. However, many manufacturing industries have vibrating hoppers,which may not be amenable to loadcell implementation. Computer vision can provide a solution by using a specific algorithm that utilizes one camera and a computer to analyze the data received from the camera. This system can indicate the fullness of the hopper. It does not require any human resources, minimizing welfare loss and the possibility of an industrial accident. We provide in this report the precise algorithm used.

\section{Related Works}

\subsection{Loadcell}

\begin{figure}[htp]
    \centering
    \includegraphics[scale = 0.25]{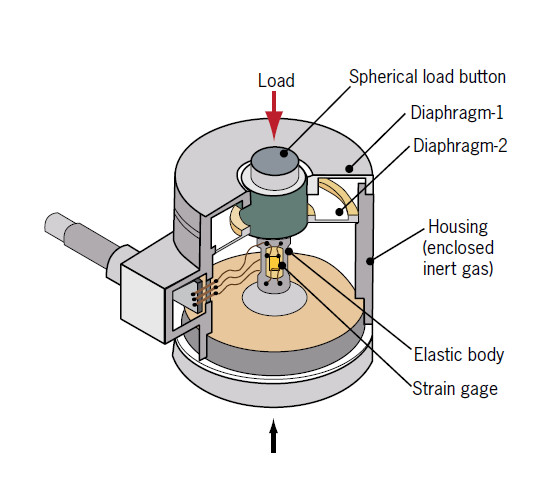}
    \caption{Simple diagram of a loadcell [6]}
    \label{fig:LoadCell}
\end{figure}

First, a load cell that measures the weight of the materials in a hopper cannot be used because most manufacturing industries use vibrating hoppers, which can equally distribute the materials inside the hopper and minimize jamming. The vibrating hopper not only wears out the load cell but could also exhibit oscillating motion that does not return a precise weight. Therefore, it is not an economically prudent solution to the problem.

\subsection{Machine Learning Techniques}

\begin{figure}[htp]
    \centering
    \includegraphics[scale = 0.25]{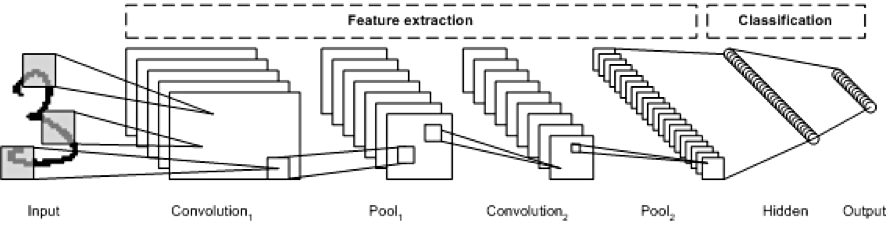}
    \caption{Simple diagram of how deep learning technique works [2]}
    \label{fig:CNND}
\end{figure}

Deep learning techniques are very useful in approaching this type of problem as they offer very precise analyses of the images and are very adept at computer vision. Nevertheless, they are not recommended as they present time lags. They do not return the results as quickly as a human eye, requiring more than 100 sample data points to achieve near-excellent learning. They also require a high-quality graphics processing unit to offer excellent data analyses, which is not necessarily profitable.

\section{Challenge and Target Scenario}

The challenge was to create an algorithm that could efficiently analyze the fullness of the hopper: 10\%, 25\%, 50\%, 75\%, or 100\%. Our first approach was to use a K means clustering algorithm to simplify the image into two colors before analysis. However, a significant outlier was identified when I used the sample data. As a result, a new algorithm was designed that uses OpenCV, standard deviation, and variance.

\subsection{Theory} The theory of the algorithm is based on the fact that more materials in the hopper affect the standard deviation of the target line from the image(target lines will be explained in upcoming subsections).

\begin{figure}[htp]
    \centering
    \includegraphics[scale = 0.085]{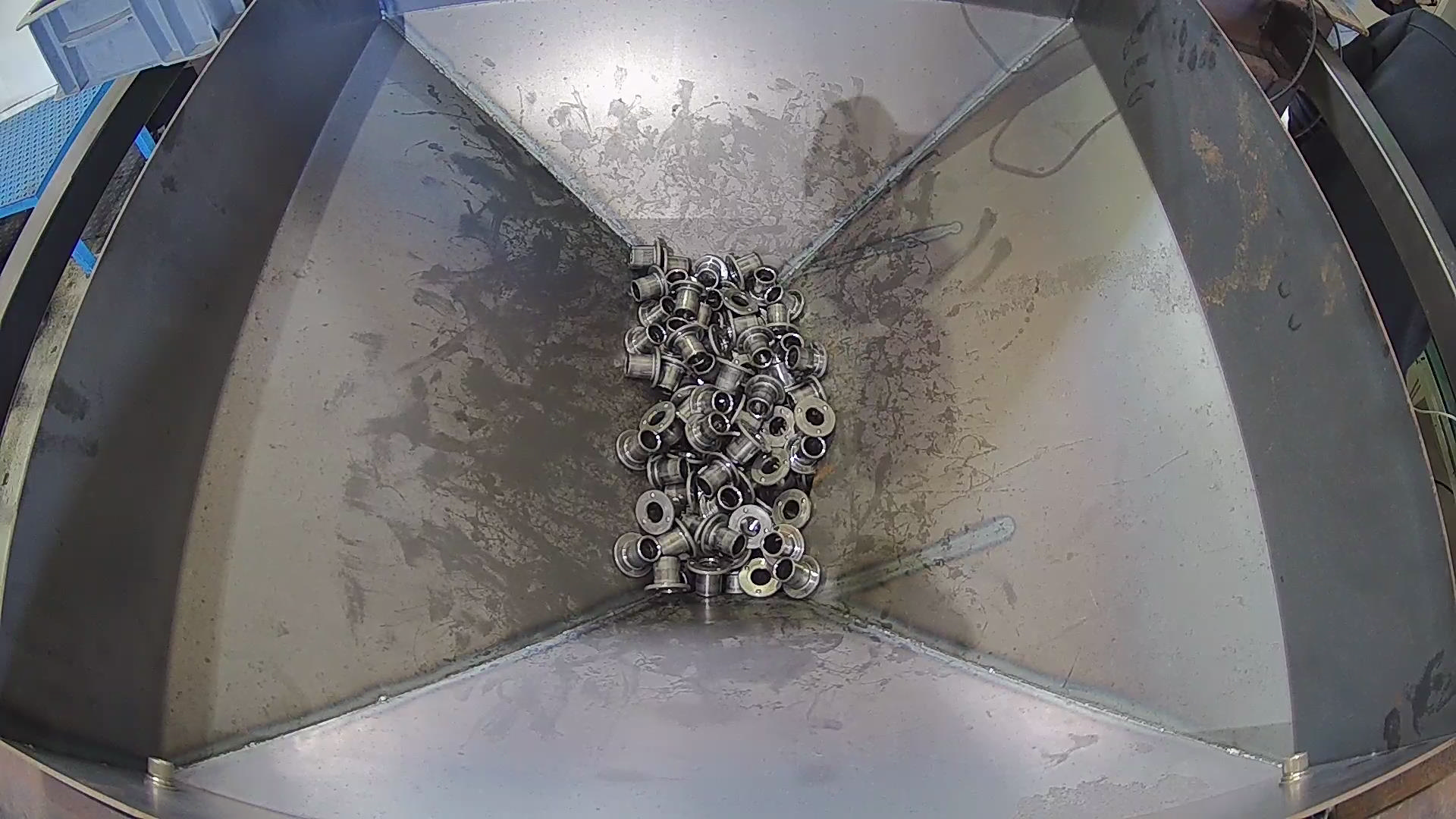}
    \caption{Sample Data of 10\% Full Hopper}
    \label{fig:sampleData}
\end{figure}

\begin{figure}[htp]
    \centering
    \includegraphics[scale = 0.14]{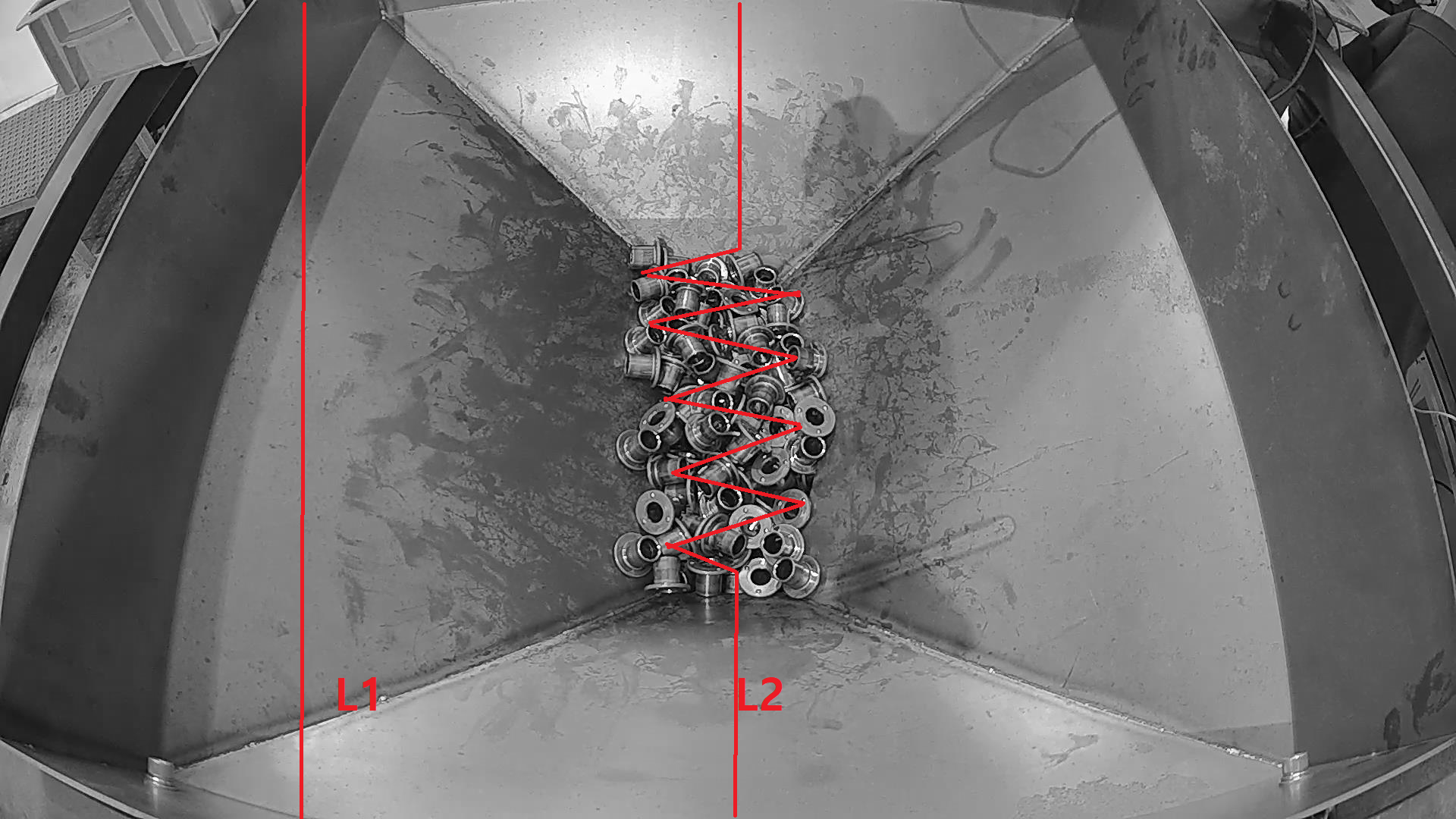}
    \caption{Modified Sample Data of 10\% Full Hopper}
    \label{fig:MsampleData10}
\end{figure}

\begin{figure}[htp]
    \centering
    \includegraphics[scale = 0.14]{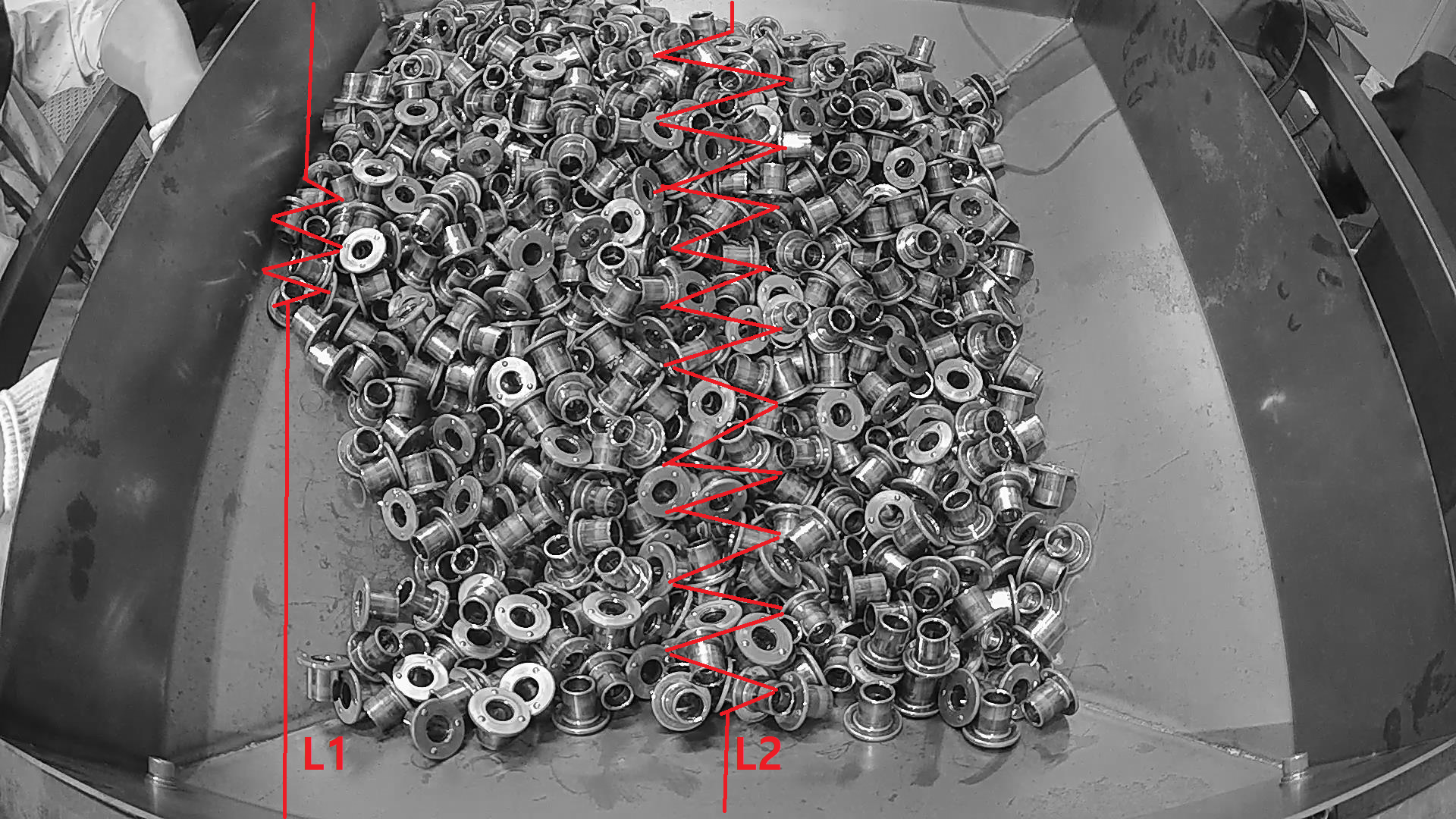}
    \caption{Modified Sample Data of 50\% Full Hopper}
    \label{fig:MsampleData50}
\end{figure}

\begin{equation}
\sigma = \sqrt{\frac{\sum (r_i-r_{avg})^2}{N}}
\end{equation}

Where $\sigma$ = standard deviation, $r_i$ = value of each pixel, $r_{avg}$ = average pixel value, and N = size of the pixel array.

\begin{equation}
\sigma^2 = \frac{\sum (r_i-r_{avg})^2}{N}
\end{equation}
Where $\sigma^2$ = variance and the rest of the variable are the same as formula number 1.

\begin{equation}
A_1 = \frac{\sigma_1 + \sigma_2}{2}
\end{equation}

Where ${\sigma_1}$ = standard deviation of pixels in L1, ${\sigma_2}$ = standard deviation of pixels in L2, and $A_1$ = arithmetic mean of the standard deviation of L1 and L2.

\begin{equation}
A_2 = \frac{\sigma_1^2 + \sigma_2^2}{2}
\end{equation}

Where ${\sigma_1^2}$ = variance of pixels in L1, ${\sigma_2^2}$ = variance of pixels in L2, and $A_2$ = arithmetic mean of the variance of L1 and L2.

Figure 4 shows sample data of a 10\% filled hopper, and Figure 5 depicts the two abstract targeted lines used for this algorithm. As seen in Figure 5, the image is converted into a greyscale image from an RGB image, and there are two lines: "L1" and "L2." L2 is vertically located at the center of the image so that L2 can determine the standard deviation at the center and determine the fullness of the hopper. The line L1 indicates whether the hopper is 75\% full or 100\% full (the materials will extend to L1 if it overflows 75\%). 

The algorithm retrieves the values of the pixels located in the lines. As the values are retrieved, the code calculates standard deviation using Formula 1. As seen in Figure 5, the standard deviation of L2 seems more affected (in terms of standard deviation) by the texture of the materials inside the hopper than that of L2 in Figure 6. The length of the zigzagged lines inside the figures depicts the degree of variation within the standard variation created by the materials inside the hopper. Had the hopper been even more filled, the zigzag portion would have been larger, increasing the standard deviation value. However, contrasting the standard deviation values revealed that the formula involving the standard deviation of L1 and L2 should be determined. 

Using the arithmetic mean of among variance, standard deviation, or $A_1^2$($A_1$ in Formula 3) was proven effective for contrasting with greater than 85\% accuracy (explained in the choice of algorithm section). Each presented pros and cons. For example, the four-digit magnitude mean of variance could be used to compare the values between each hopper type. The squared arithmetic mean of the standard deviation also exhibited four-digit magnitudes that provided the same benefits as the mean of variance but with different values. For the arithmetic mean, however, the magnitudes were not large enough(two-digit) to determine the precise threshold between each percentage of hoppers which can be problematic, and it is extremely similar to the squared arithmetic mean. In fact, the larger the magnitude, the greater the range of errors permitted. Therefore, the choice of algorithm was narrowed to the mean of variance and the squared value of the arithmetic mean of standard deviation. 

As there are different types and designs of hoppers in industry and smart factories, the standard deviation of the empty hopper image could be subtracted from the returned value. The different hopper designs can reflect different textures. The amount of rust or stain will also differ, which may affect the pixel value of the image, leading to an inaccurate standard deviation value. When subtracted, however, the result provides the standard deviation of the texture of the materials inside the hopper. In that case, the range in which the standard deviation could indicate 10\%, 25\%, 50\%,75\%, or 100\% would theoretically be the same as all the other hoppers. Nevertheless, the precise experimental results could not be produced since there were fewer images of empty hoppers in modern smart factory industries; thus, this could be considered an inference.

\subsection{Experimental Results of Algorithms}

\begin{figure}[htp]
    \centering
    \includegraphics[scale = 0.67]{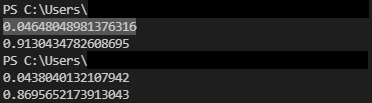}
    \caption{The Accuracy of the Algorithm}
    \label{fig:Result}
\end{figure}

As seen in Figure 7, the accuracy of the algorithms tended to be more than 85\%, and the average time taken for the code to analyze the image was less than a second, substantially less than 1.8s of run time and slightly over 80\% accuracy for existing deep learning techniques in the common smart factory industry. However, it may be debatable whether 0.003s of run time is worth 5\% accuracy compared to each algorithm mentioned. This question may not be debatable when only seeing the 5\% value, but only 23 data points were given. 1/23 is approximately 4.3\%, which is related to the difference in accuracy in that the accuracy of the algorithm differs only by one test case. As mentioned regarding the vibrating hopper in the first section, the given test case images were collected in extreme environments (such as in Figure 6, where the materials are skewed to one side). Therefore, the mentioned algorithms may exhibit high accuracy in the real data set. Specific explanations will be included in the evaluation section.

\section{Evaluation}

The overall algorithm seems useful, and we will now describe the process used to prove the effectiveness of the algorithm. 

\subsection{Run Time Measurement and Accuracy Measurement}

\begin{figure}[htp]
    \centering
    \includegraphics[scale = 1]{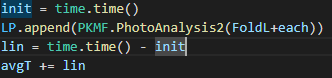}
    \caption{Sample code to determine the run-time}
    \label{fig:runTimeSampleCode}
\end{figure}

In Figure 7, the algorithm's run-time and accuracy can be seen. The approach to measuring the run time was using Python's "time" library. It is a general approach to determine the run-time of the code but necessary to determine the run-time for 23 data points at once. Thus, I used the "time" library to determine the run-time of each of the 23 test data points and averaged them (Figure 7). The accuracy also considers all the outliers (an inaccurate output may be considered to have an algorithm return of 50\% instead of 75\% when the actual image is 75\%) as wrong outputs, so it only returns
\begin{center}
$R/T = accuracy$ 
\end{center}

where 

\begin{center}
$R = $ Correct number of outputs 
\end{center}

and 

\begin{center}
$T = $ Total number of outputs.
\end{center}

Figure 8 shows how it was coded to determine the run-time for the code. As a practical example, FronTech Inc., located in the Republic of Korea, runs 50 machine hoppers in total. Had the algorithm been based on deep learning techniques, a server computer would require a total of 60 seconds to process the images from 50 cameras (1.2s for an image to be processed * 50 = 60s). In a FronTech Inc. industrial operation, a worker can check 9~10 hoppers in one minute. However, the suggested algorithm requires only 2.3 seconds to process 50 data points from the cameras(0.046 for an image to be processed * 50 = 2.3s). Indeed, the company would require at least 5 GPU-installed computers to process each camera using deep learning techniques. However, the suggested algorithm requires only 1 for 50, which is economically and technically efficient.

\subsection{Test Data Sets and Definitions}

\begin{figure}[htp]
    \centering
    \includegraphics[scale = 0.085]{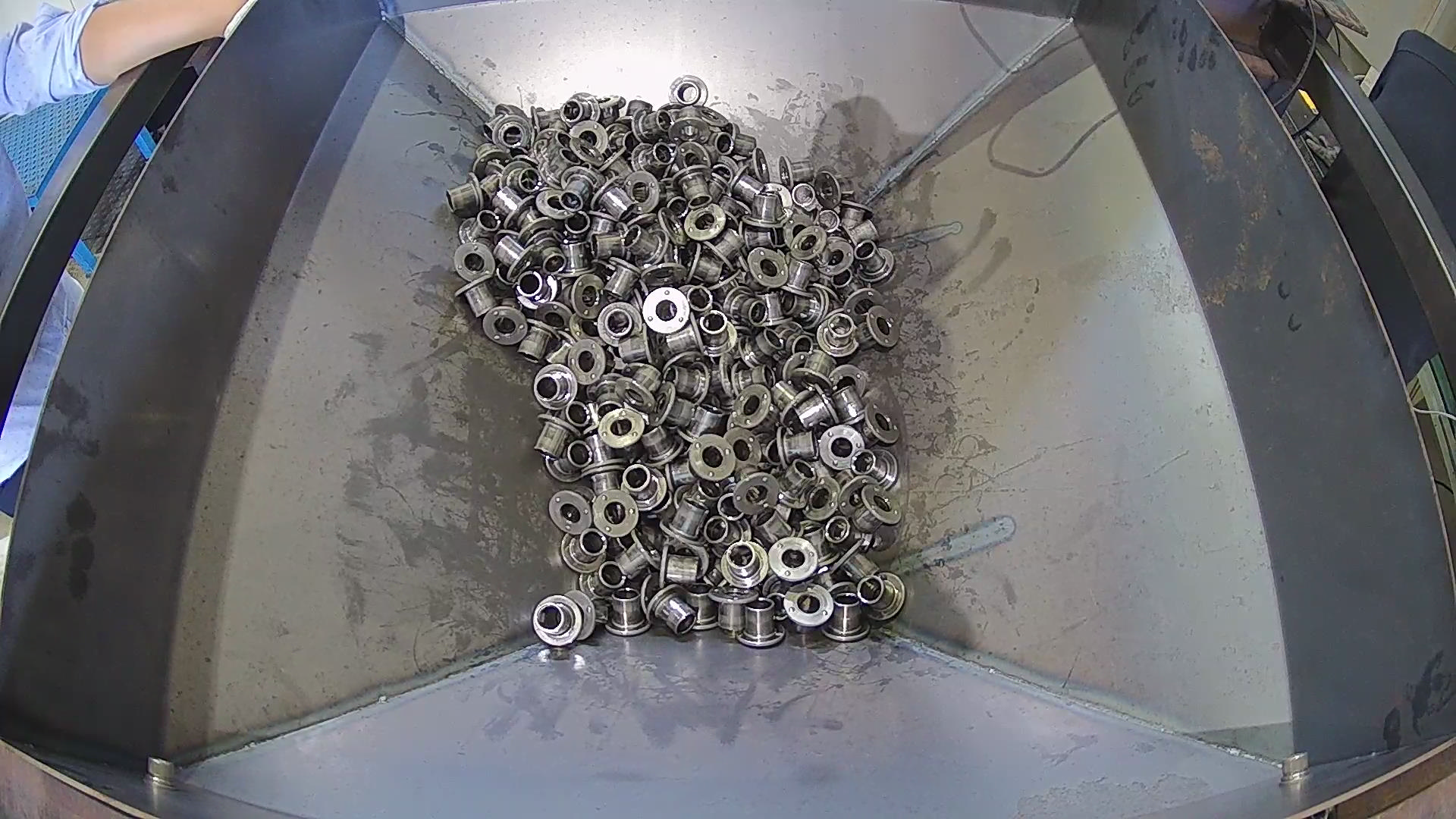}
    \caption{Test Data of 25\% Full Hopper}
    \label{fig:skewedLeftTop}
\end{figure}

\begin{figure}[htp]
    \centering
    \includegraphics[scale = 0.085]{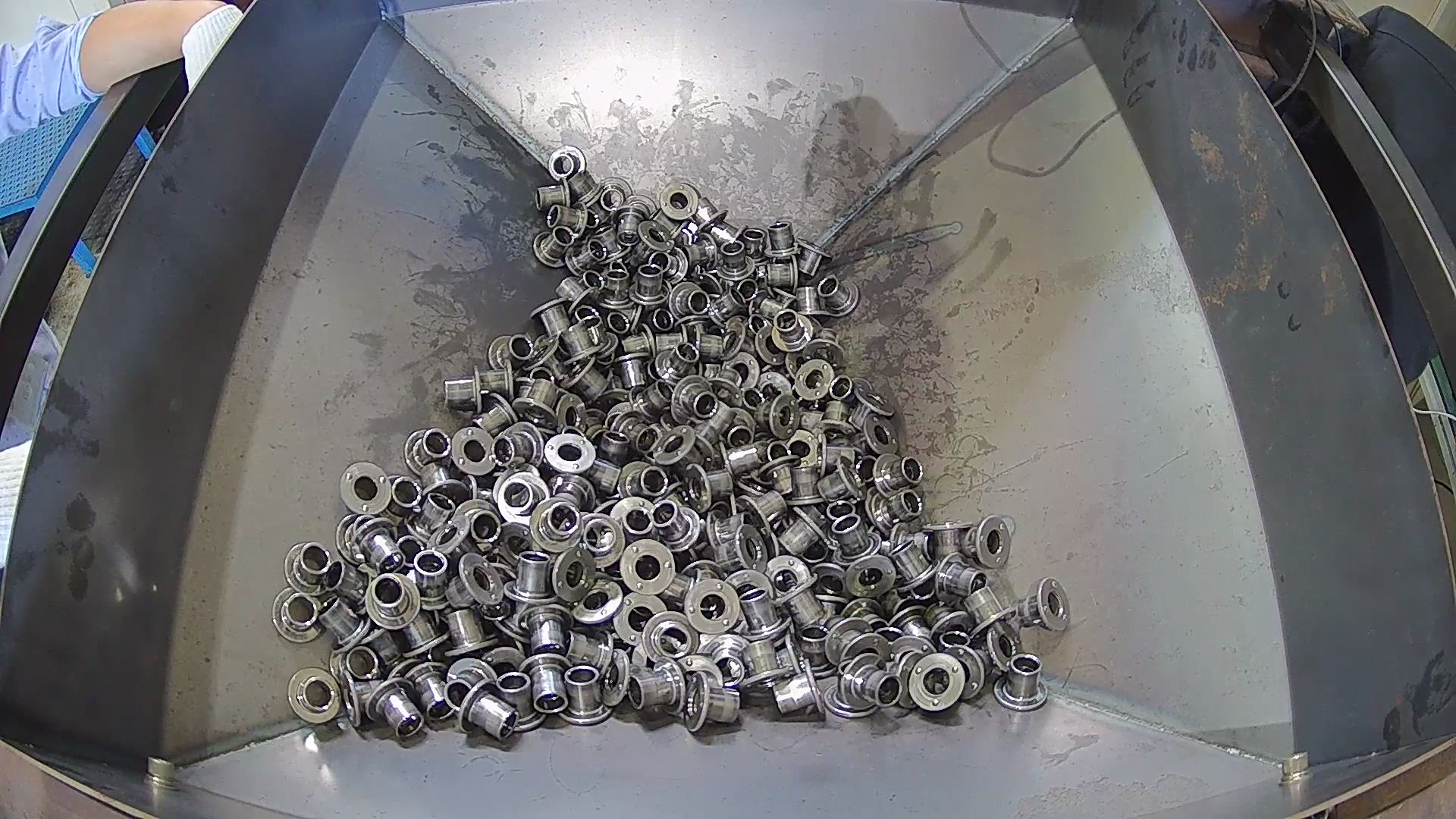}
    \caption{Test Data of 25\% Full Hopper}
    \label{fig:skewedLeftBottom}
\end{figure}

Only 23 test data points were provided to verify the algorithm. In addition, most of the data sets were recorded in extreme cases. For example, Figure 9 shows skewed material in the left top corner, which is biased data and highly unlikely for a vibrating hopper. Figure 10 shows biased data where the materials are skewed towards the left bottom. Had the hopper been vibrating, the materials would likely be nearly equally distributed due to gravitational law.

\begin{figure}[htp]
    \centering
    \includegraphics[scale = 0.085]{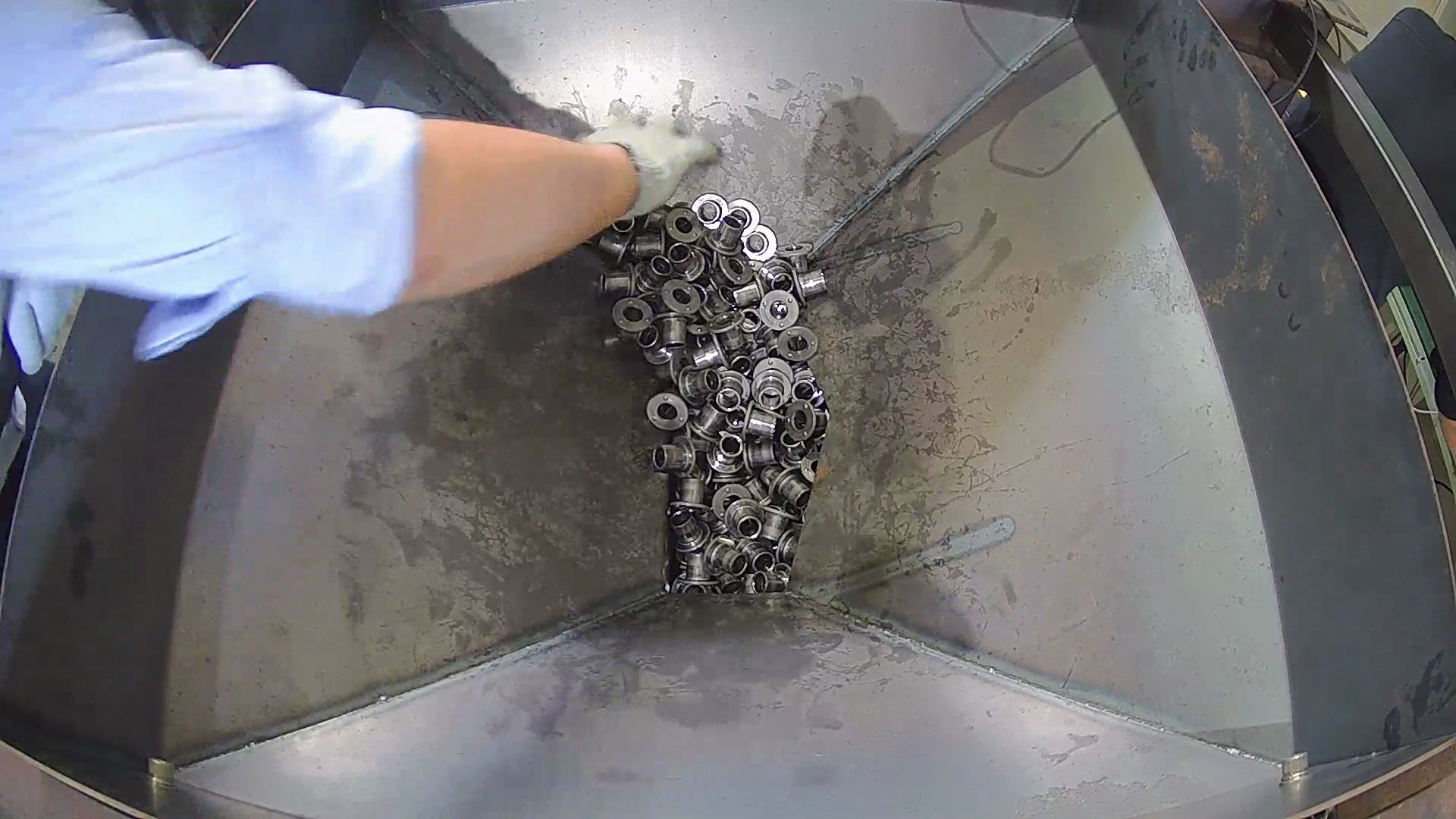}
    \caption{Test Data of 10\% Full Hopper}
    \label{fig:improperData}
\end{figure}

\begin{figure}[htp]
    \centering
    \includegraphics[scale = 0.085]{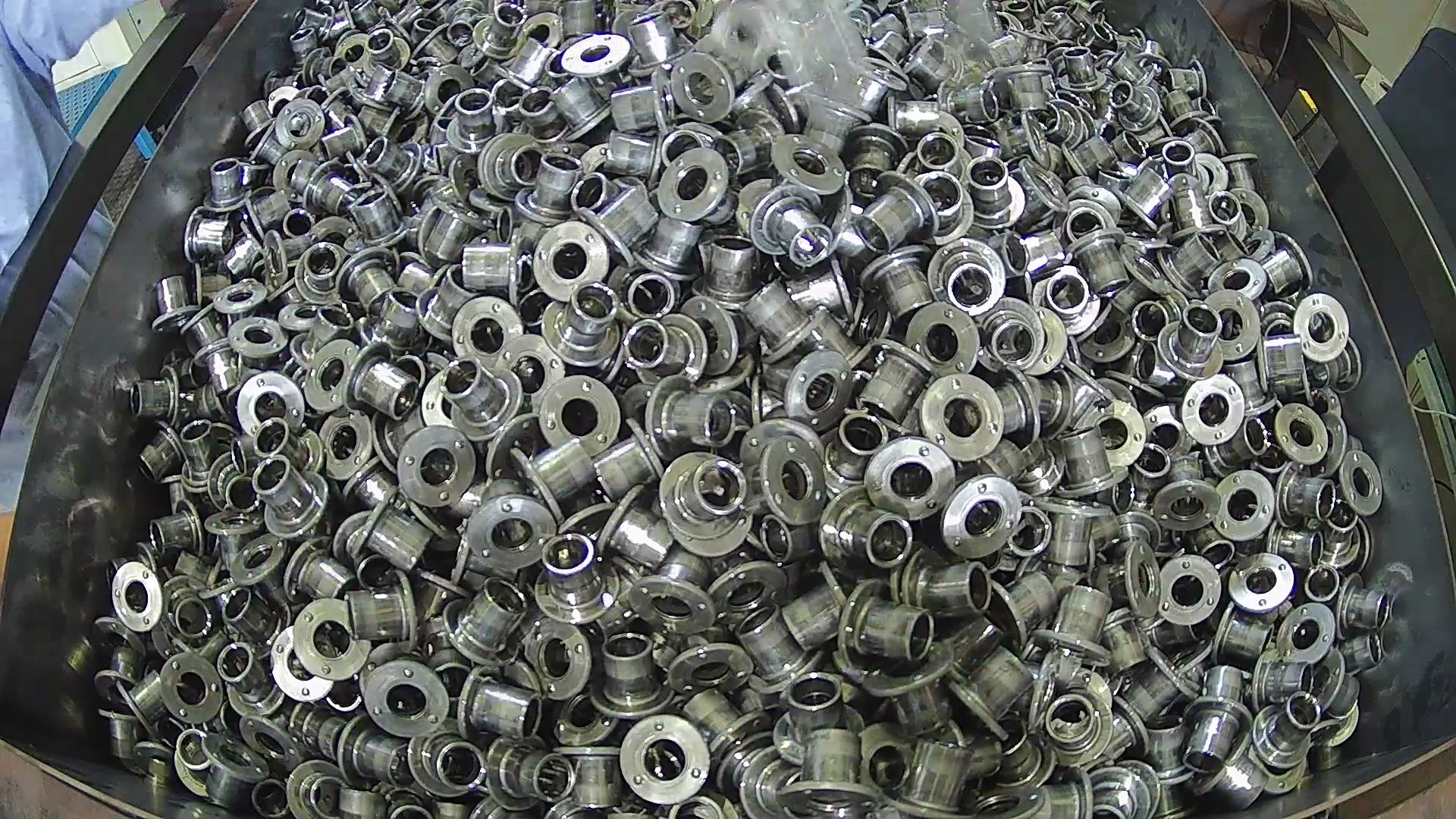}
    \caption{Test Data of 100\% Full Hopper}
    \label{fig:improperData2}
\end{figure}

Some data, however, were not appropriate. Figure 11 shows a 10\% filled hopper, but the worker's arm is observed, which would result in an improper output. Figure 12 also shows data where the picture was taken while the objects were falling. As a result, 2 of 25 data were inappropriate and had to be removed. 

\begin{figure}[htp]
    \centering
    \includegraphics[scale = 0.085]{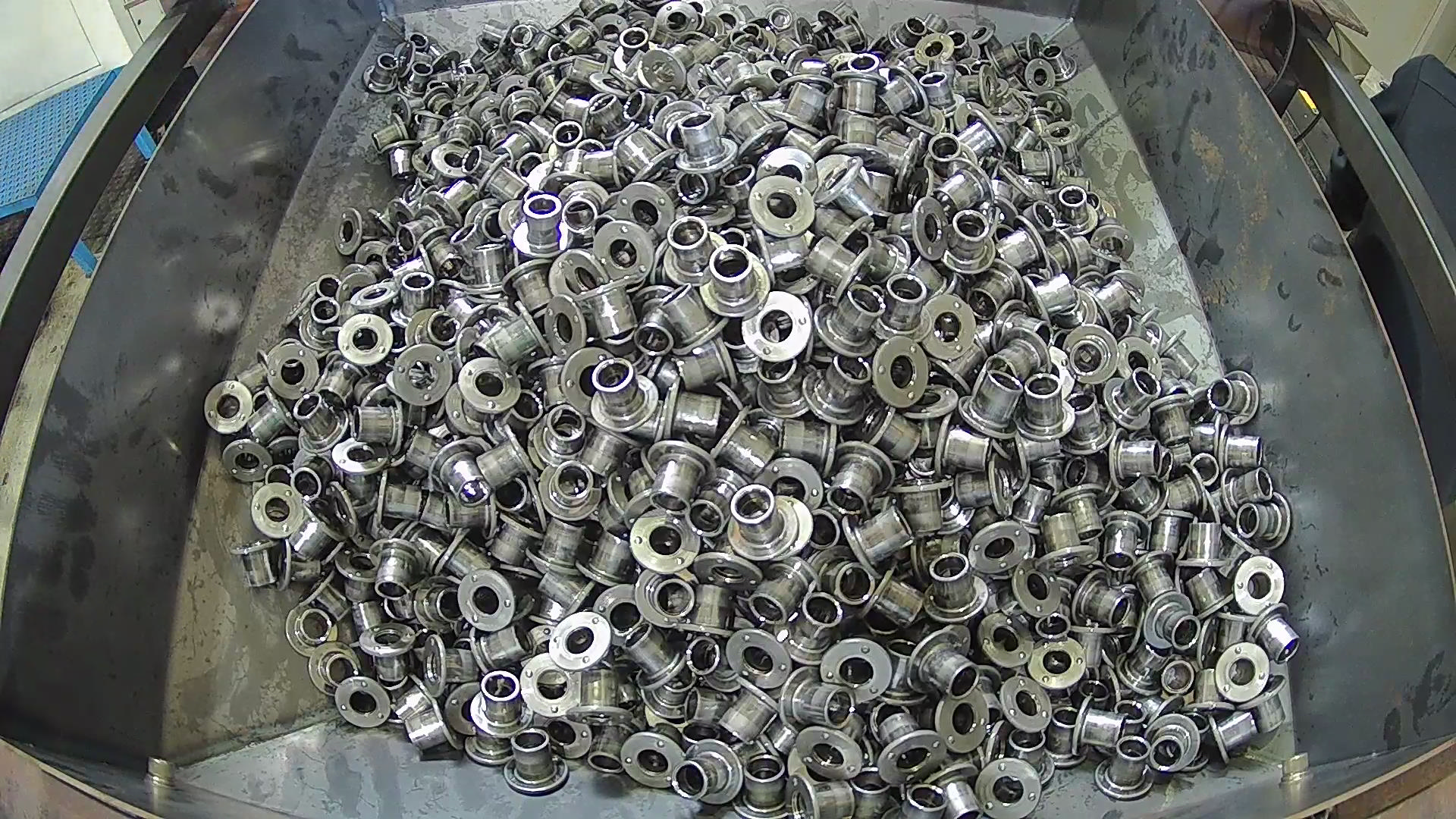}
    \caption{Test Data of 75\% Full Hopper}
    \label{fig:incorrectOutput}
\end{figure}

The algorithm could be validated if there were at least 50 appropriate data points. Currently, however, there is no alternative way to validate the data. Though the data is biased, using the given data would be the only data available to validate my algorithm. Figure 13 shows one of the data points that the algorithm could not correctly identify. FronTech Inc. stated that the data exhibits a 75\% filled hopper, but the algorithm identifies the hopper as 50\% full. Visually, people would likely identify the data as a 50\%-filled hopper. They are stacked on top of each other; as the hopper vibrates, however, they should be 
distributed equally.

Not only the data set, but how FronTech Inc. defined 10\%, 25\%, 50\%, 75\%, and 100\% was abstract. There is likely a general tendency for patterns to occur, but not all the data could be reported in exact percentages. In fact, data such as that shown in Figure 13 may be problematic because deciding the range of standard deviation for each percentage may force the user to create a more incorrect range. Data such as those in Figure 13 could also be considered outliers.

\section{Future Work \& Discussion}

The techniques mentioned can be used to monitor hoppers in various smart factory industries. Certain factors will require manipulation to keep track of the materials correctly, such as the coordinates of the L1 and L2 line axes. The algorithm could be modified to identify the coordinates of L1 and L2 automatically from the hopper image, increase the number of lines, and count the number of specific materials in the hopper. 

The suggested algorithm may be viewed as simple. Nonetheless, such algorithms are not investigated because domestic firms in many countries demand deep learning techniques and luxury algorithms that could serve to advertise their companies. These advanced solutions to problems such as the one addressed here do not improve welfare and are inefficient. In fact, sometimes, a simpler solution is necessary.

\section{Conclusion}

Overall, it was a lengthy process to create the algorithm and correct the data. Though there were limitations during algorithm validation, the algorithm itself still exhibits high accuracy and seems readily deployable in real-world industrial situations. It was challenging to choose the specific method between the arithmetic mean of variance and the square of average standard deviation between L1 and L2. In determining which algorithm to sacrifice, I decided to include both in the GitHub link[4]. The algorithm may have more potential in the computer vision field, and it could become an algorithm that stands as an alternative to deep learning techniques. It requires refinement, and outliers are present as it relies on the texture of the materials. It is likely applicable to solutions to many other challenges. %{https://github.com/seunghk1206/Residual-Quantity-in-Percentage-of-Factory-Machines-Using-Computer-Vision-and-Mathematical-Methods}

%%%%%%%%%%%%%%%%%%%%%%%%%%%%%%%%%%%%%%%%%%%%%%%%%%%%%%%%%%%%%%%%%%%%%%%%%%%%%%%%

%%%%%%%%%%%%%%%%%%%%%%%%%%%%%%%%%%%%%%%%%%%%%%%%%%%%%%%%%%%%%%%%%%%%%%%%%%%%%%%%


\begin{thebibliography}{99}

\bibitem{c1} Amilkanthawar, Vivek. “Choosing a Deep Learning Framework.” Medium, The HumAIn Blog, 28 Feb. 2019, https://medium.com/in-pursuit-of-artificial-intelligence/choosing-a-deep-learning-framework-5669a85ebc3f. 
\bibitem{c2} “CNN, Convolutional Neural Network abstract.” Blog main page, 4 Jan. 2018, http://taewan.kim/post/cnn/. 
\bibitem{c3} “Home.” Magnetic Food Feeding Vibratory Hopper - Buy Magnetic Vibratory Feeder,Food Feeding Conveyor,Food Industry Vibrating Feeder Product on Alibaba.com, https://www.alibaba.com/product-detail/Magnetic-food-feeding-vibratory-hopper\_60488056903.html. 
\bibitem{c4} Kim, Seunghyeon. “Residual Quantity in Percentage of Factory Machines Using Computer Vision and Mathematical Methods.” GitHub, 1 Nov. 2021, https://github.com/seunghk1206/Residual-Quantity-in-Percentage-of-Factory-Machines-Using-Computer-Vision-and-Mathematical-Methods/settings. 
\bibitem{c5} “Standard Deviation.” Corporate Finance Institute, 31 Jan. 2020, https://corporatefinanceinstitute.com/resources/knowledge/standard-deviation/. 
\bibitem{c6} WhitePaper. “Load Cell vs. Force Sensor.” Tekscan, 5 June 2020, https://www.tekscan.com/resources/whitepaper/load-cell-vs-force-sensor. 
\end{thebibliography}
\end{document}